\documentclass{article}

\usepackage[final]{corl_2026} %

\usepackage{amssymb}
\usepackage{graphics} %
\usepackage{epsfig} %
\usepackage{times} %
\usepackage{pifont}
\usepackage{amsmath} %
\usepackage[dvipsnames]{xcolor}
\usepackage{xspace}
\usepackage{booktabs}
\usepackage{wrapfig}
\usepackage{enumitem}
\usepackage{algorithm}
\usepackage{algpseudocode}
\usepackage{subcaption}
\usepackage{float}

\definecolor{idmblue}{HTML}{1F4E79}      %
\definecolor{polorange}{HTML}{B7472A}    %
\definecolor{aligngreen}{HTML}{2E7D32}   %

\makeatletter
\DeclareRobustCommand\onedot{\futurelet\@let@token\@onedot}
\def\@onedot{\ifx\@let@token.\else.\null\fi\xspace}

\def\eg{\emph{e.g}\onedot} 
\def\ie{\emph{i.e}\onedot}

\makeatother

\newcommand{\methodname}[1]{\textsc{Ela}}

\title{ELA: End-to-End Latent Action Policy Learning from Video Demonstrations}

\author{
  Cheng-Chun Hsu \quad
  Peter Stone \quad \\
  Department of Computer Science, The University of Texas at Austin \qquad \\
  {\texttt{\{chengchun, pstone\}@utexas.edu}} \qquad \\
  \\[0.5em]
  \href{https://end2end-latent-action.github.io/}{end2end-latent-action.github.io}
}

\begin{document}
\maketitle

\begin{abstract}
Robot imitation learning is limited by the scarcity of action-labeled data, while action-free demonstrations such as videos are abundant. Recent latent-action methods learn action-relevant representations from observation transitions, but they do not optimize latent actions for downstream control, leading to latent representations that are difficult for policies to model and difficult to accurately decode into continuous robot actions.
This paper introduces \methodname{}, an end-to-end\footnote{We use ``end-to-end'' to indicate that all components are optimized toward the downstream objective, without meaning direct gradient propagation through all modules.} framework for learning from video demonstrations via latent actions jointly optimized for policy learning and action decoding. \methodname{} jointly trains a latent action model, latent action policy, and action decoder, enabling policies to leverage abundant video data with only a handful of action-labeled robot demonstrations.
Experiments on simulated and real-world manipulation tasks show that \methodname{} consistently outperforms prior latent action baselines. Moreover, it integrates seamlessly with existing imitation learning methods through lightweight 4M-parameter action decoders, requiring no changes to policy architectures or training objectives.
Additional materials can be found on: \href{https://end2end-latent-action.github.io/}{end2end-latent-action.github.io}

\end{abstract}

\keywords{Latent Action Modeling, Learning from Videos}

\begin{figure}[h]
    \centering
    \vspace{-13pt}
    \includegraphics[width=1.0\textwidth]{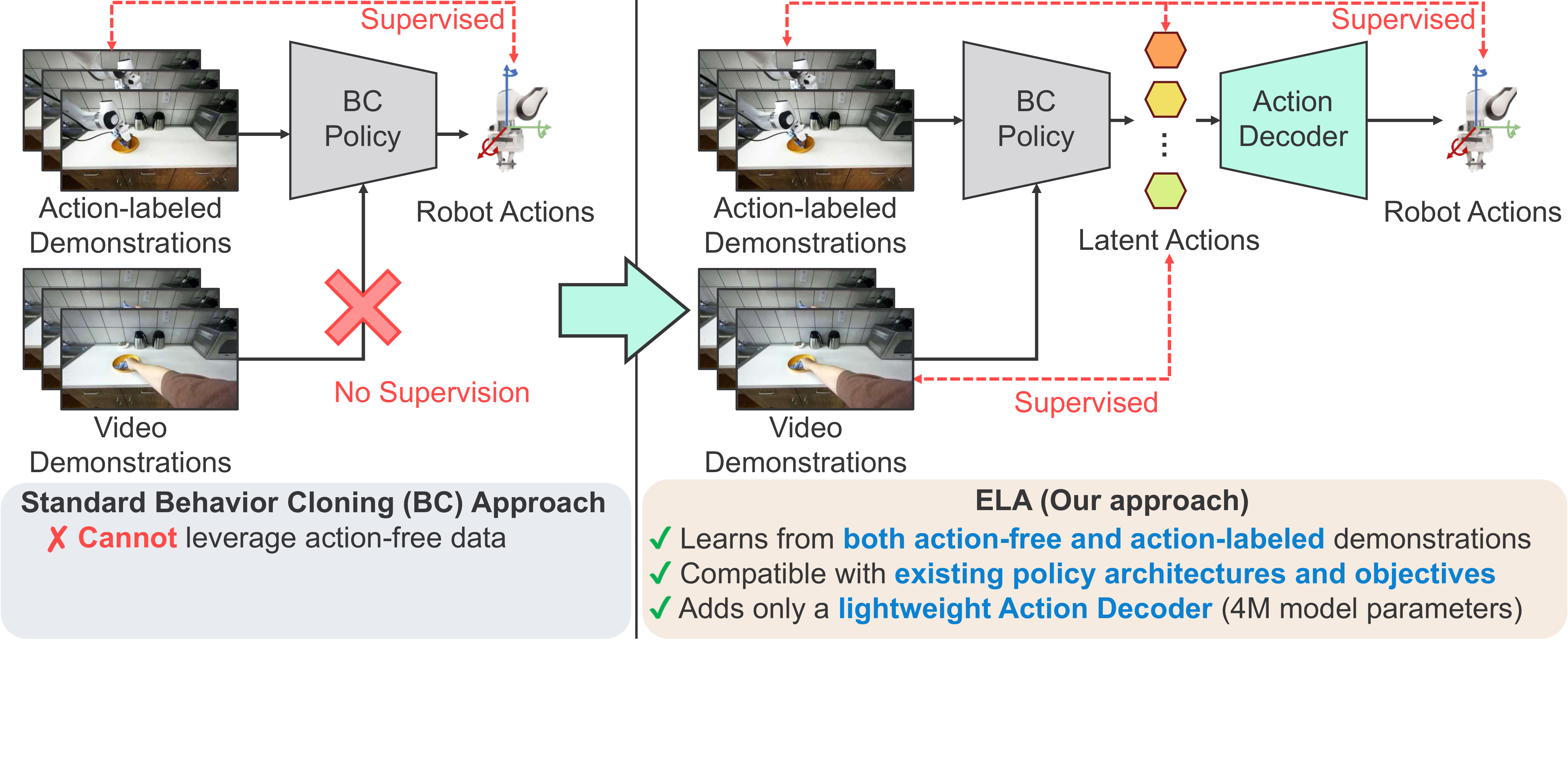}
    \vspace{-52pt}
    \caption{
        \textbf{\methodname{} Overview.} Our framework extends existing behavior cloning methods,~\eg diffusion policies, with latent actions jointly optimized for policy learning and action decoding. The approach enables training on mixtures of action-labeled and video demonstrations with little overhead, requiring only a lightweight action decoder.
    }
    \label{fig: teaser}
    \vspace{-15pt}
\end{figure}

\section{Introduction}
Data scarcity remains a fundamental bottleneck for robot imitation learning, because collecting large-scale datasets~\cite{o2024open, khazatsky2024droid} with robot action labels is expensive and often infeasible. In contrast, video demonstrations, including human videos and unlabeled robot trajectories, are significantly easier to obtain at scale. This contrast motivates imitation learning approaches that learn directly from video data without requiring action labels.

Latent action modeling~\cite{schmidt2024learning, ye2025latent} has recently emerged as a promising direction for leveraging such data. These methods infer action-relevant structure from video demonstrations and use it as the supervision target for policy learning, removing the need for explicit action supervision. This property makes latent action learning particularly promising for leveraging large-scale video data for imitation learning. However, while effective as an abstract representation, existing approaches do not explicitly optimize the latent action space for downstream control.

We identify two key properties that a latent action space should satisfy for effective policy learning: \emph{modelability}, meaning latent actions can be reliably inferred from observations for policy learning, and \emph{decodability}, meaning they can be faithfully mapped into executable continuous robot actions. Existing methods~\cite{ye2025latent, kim2025uniskill, bu2025univla, routray2025vipra} typically adopt a two-stage pipeline, where latent actions are first learned from image pairs and then used as fixed targets for a downstream policy. This decoupling leads to a latent space that is not explicitly shaped for either reliable inference from observations or accurate execution, potentially resulting in latent actions that are difficult to predict or decode.

In this work, we study whether latent actions can instead serve as a unified interface for imitation learning that is jointly optimized for both \emph{modelability} and \emph{decodability}. To that end, this paper introduces the \methodname{} framework shown in Figure~\ref{fig: teaser}, in which latent actions act as both the supervision signal for policy learning and the interface for action execution. By jointly optimizing the latent space with respect to both policy learning and action decoding, our method learns latent actions that can effectively leverage action-free video data while requiring relatively little action-labeled data for decoding into continuous robot actions.

\methodname{} learns from a mixture of action-labeled and action-free demonstrations by jointly training three components: a Latent Action Model, a Latent Action Policy, and an Action Decoder. The Latent Action Model infers a latent action responsible for each state transition, which are then used as supervision for the Latent Action Policy. This approach enables policy learning directly from video data without modifying the underlying policy architecture or objective, while ensuring \emph{modelability} by shaping a latent space such that latent actions corresponding to demonstrations are predictable. The Action Decoder maps latent actions into continuous robot control commands, ensuring \emph{decodability}. To prevent latent action collapse during joint training, we further introduce a representation alignment objective that stabilizes the latent space.

Overall, \methodname{} enables imitation learning from video demonstrations by learning a latent action space that is explicitly optimized for both \emph{modelability} and \emph{decodability}. It effectively leverages video data and outperforms other latent action baselines across simulated and real-world benchmarks. Moreover, it can be integrated into existing imitation learning approaches without modifying policy architectures or training objectives, using a lightweight Action Decoder that adds a fixed 4M-parameter overhead independent of policy size.

\section{Related Work}
Learning from video demonstrations~\cite{torabi2018behavioral} enables robots to acquire skills without requiring action labels, alleviating the data scarcity bottleneck in robot imitation learning.
Prior works extracts action-relevant information from videos through intermediate representations such as hand poses~\cite{haldar2025point, hong2025hand}, object poses~\cite{hsu2025spot, su2025motion}, and point tracks~\cite{wen2023any, collins2025amplify}. These methods typically rely on external vision systems, including hand pose predictors~\cite{chen2025large} and object pose estimators~\cite{patel2025robotic}, and often require carefully designed architectures tailored to specific representations.
Another line of work, world action models (WAM)~\cite{zhu2025unified, li2025unified, pai2025mimic}, leverages video data through world modeling objectives such as next-frame prediction. While effective for learning visual dynamics, these approaches generally do not recover action representations that can directly supervise policy learning.

Latent action modeling~\cite{schmidt2024learning, ye2025latent} instead learns inverse dynamics models (IDMs) that infer latent actions from video, enabling training on diverse data sources including human videos~\cite{yang2025egovla} and robot datasets~\cite{khazatsky2024droid}. However, latent actions are often too abstract for direct low-level control and are mainly used for policy pretraining~\cite{ye2025latent, kim2025uniskill, bu2025univla, routray2025vipra}.
Recent methods~\cite{liang2025clam, chen2025villa, zhang2026clap} incorporate proprioception and action supervision, but the separate training of the IDM and policy often leads to sub-optimal performance. In contrast, \methodname{} jointly learns the inverse dynamics model and control policy, enabling direct end-to-end policy learning from action-labeled and action-free demonstrations.

\section{Approach}

\begin{figure*}[t]
    \centering
    \includegraphics[width=1.0\textwidth]{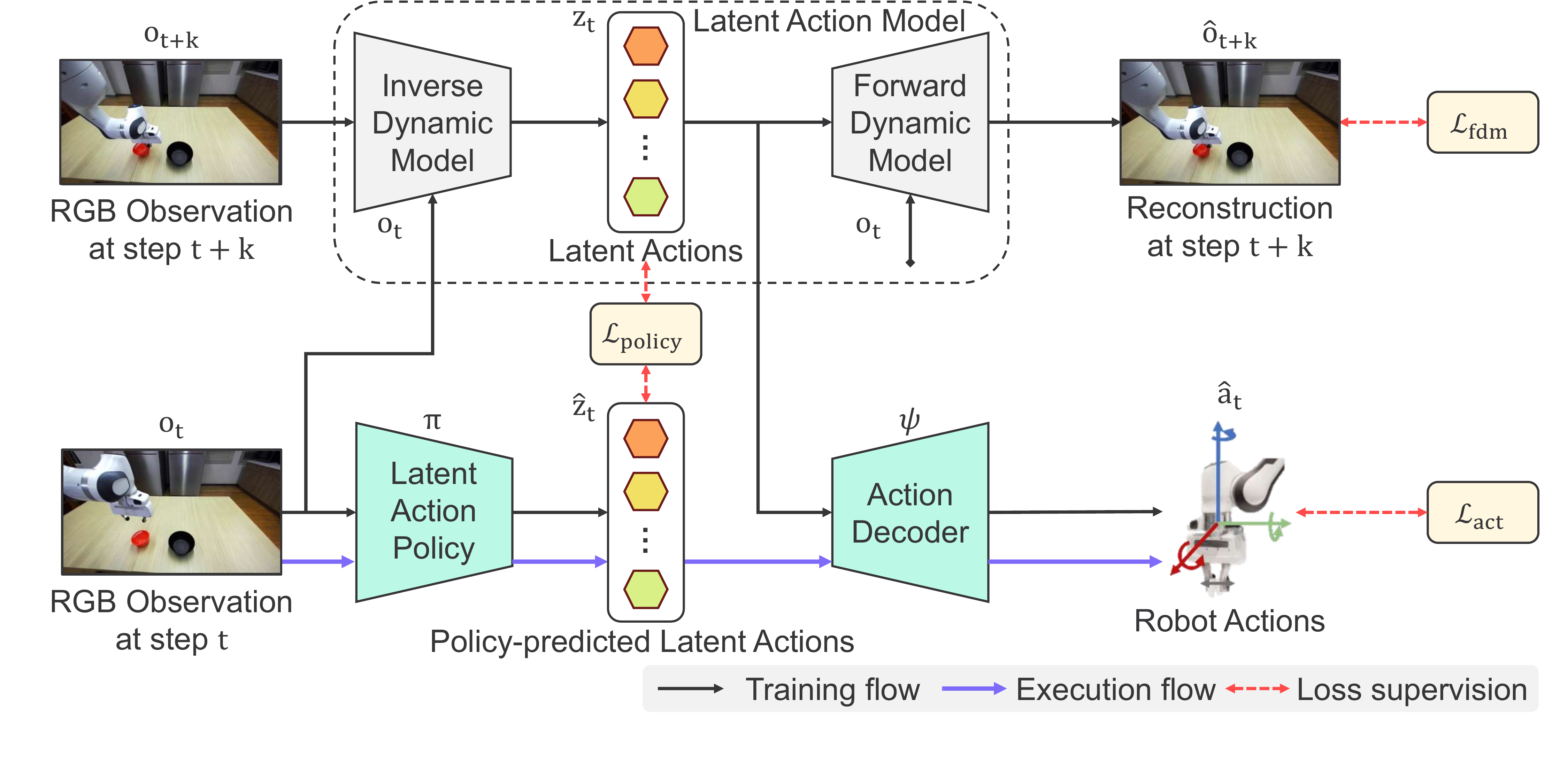}
    \vspace{-35pt}
    \caption{
        \textbf{\methodname{} Training and Deployment.}
        During training, image pairs are used with an inverse dynamics model (IDM) to infer latent actions, which supervise a forward dynamics model (FDM) for future frame reconstruction and an action decoder for robot action prediction. A latent action policy is then trained to output the latent actions.
        During deployment, the latent action policy maps image observations to latent actions, which are decoded into robot actions by the action decoder.
    }
    \label{fig: framework}
    \vspace{-20pt}
\end{figure*}

In this section, we introduce \methodname{}, a latent-action approach for learning from a mixture of action-labeled and action-free demonstrations. \methodname{} jointly trains three components, a Latent Action Model, a Latent Action Policy, and an Action Decoder. A high-level overview of the proposed architecture is shown in Fig.~\ref{fig: framework}.

\subsection{Problem Formulation}
We assume access to two types of expert demonstration data: an action-free dataset of RGB observation trajectories $\mathcal{V} = \{o_t\}$ and a smaller action-labeled dataset $\mathcal{R} = \{(o_t, a_t)\}$, where $o_t \in \mathcal{O}$ denotes RGB image observations at time $t$ and $a_t \in \mathcal{A}$ denotes expert robot actions at time $t$. The action-free dataset may come from the same embodiment as the target robot or from a different source,~\eg, human demonstrations, whereas the action-labeled dataset is collected on the target embodiment and provides ground-truth control commands.
Given these datasets, our objective is to learn a vision-based imitation learning policy $\Pi(a_t \mid o_t)$ that executes robot actions $a_t$ from observations $o_t$ to achieve task success, measured by a binary success signal.

\subsection{Method Overview}
\label{sec: method-overview}

To learn from both action-free and action-labeled data, we first introduce a latent action model consisting of an inverse dynamics model ($\text{IDM}$) and a forward dynamics model ($\text{FDM}$). For policy learning, we use the $\text{IDM}$, which maps a pair of observations separated by $k$ time steps to a latent action: $\mathbf{z}_t = \text{IDM}(o_t, o_{t+k})$. Next, we decompose the policy into a latent action policy and an action decoder: $\Pi(a_{t:t+k} \mid o_t) = \psi(a_{t:t+k} \mid \mathbf{z}_t)\,\pi(\mathbf{z}_t \mid o_t)$, where the latent action policy $\pi: \mathcal{O} \rightarrow \mathcal{Z}$ outputs a latent action $\mathbf{z}_t$ from a single observation $o_t$, and the action decoder $\psi: \mathcal{Z} \rightarrow \mathcal{A}^k$ maps each latent action to a $k$-step sequence of robot actions. This decoupling allows the latent action policy $\pi$ to be trained using both action-free and action-labeled data from datasets $\mathcal{V}$ and $\mathcal{R}$, since latent actions are inferred by the IDM. In contrast, the action decoder $\psi$ is trained exclusively on action-labeled data from $\mathcal{R}$.

The latent action serves as the key intermediate representation connecting the policy and the action decoder. It acts both as the prediction target for the behavior cloning policy $\pi$ and as the sole input to the action decoder $\psi$. To learn this representation effectively, we introduce \methodname{}, which jointly trains three components: the inverse dynamics model $\text{IDM}$, the latent action policy $\pi$, and the action decoder $\psi$, in order to shape the latent action $\mathbf{z}$.

\subsection{Latent Action Learning}
We instantiate latent actions by modeling state transitions over $k$ time steps. Given observation pairs $(o_t, o_{t+k})$, we learn a latent action $z_t$ that explains the transition dynamics induced by human or robot. To learn this latent action, we parameterize a latent action model consisting of an inverse dynamics model (IDM), $p_{\text{IDM}}(z_t \mid o_t, o_{t+k})$, which infers latent actions from observed transitions, and a forward dynamics model (FDM), $p_{\text{FDM}}(o_{t+k} \mid o_t, z_t)$, which enforces that the latent action is sufficient to predict future observations.

The learned latent actions serve as the interface between policy and action decoder, enabling both \emph{modelability} and \emph{decodability}. \emph{Modelability} is realized by learning a latent policy $\pi(\mathbf{z}_t \mid o_t)$ that infers latent actions from observations, while \emph{decodability} is ensured by an action decoder $\psi(a_{t:t+k} \mid \mathbf{z}_t)$ that maps latent actions to continuous robot actions.

\noindent\textbf{Inverse and Forward Dynamics Model.}
We implement the IDM by tokenizing $(o_t, o_{t+k})$ and processing them with a transformer encoder and learnable register tokens~\cite{darcet2023vision}, which aggregate inter-frame dynamics into a fixed-length latent action. A possible failure mode is that the IDM captures action-irrelevant details, or even memorizes raw image content due to the reconstruction objective. To prevent these possibilities, prior work~\cite{ye2025latent} applies quantization as an information bottleneck, discretizing latent actions into tokens so that the learned representations are encouraged to retain only action-relevant information. Following this idea, we apply an information regularization scheme based on iFSQ~\cite{lin2026ifsq} quantization. The IDM and FDM are jointly trained using an image reconstruction loss $\mathcal{L}_{\text{fdm}} = \left\| \hat{o}_{t+k} - o_{t+k} \right\|_2^2$, where $\hat{o}_{t+k} = p_{\text{FDM}}(o_t, z_t)$.

\noindent\textbf{Latent Action Policy.}
We build our latent action policy by adapting existing behavior cloning approaches~\cite{chi2025diffusion, zhu2025unified} to operate in the latent action space. Instead of predicting low-level robot actions directly, the policy $\pi$ takes an observation $o_t$ as input and outputs a latent action $z_t$. This latent action prediction is then passed to the action decoder $\psi$ to obtain executable robot actions. This modularity allows standard policy architectures to learn from latent action supervision without requiring changes to their underlying structure or training objectives. The policy is trained by the policy loss $\mathcal{L}_{\text{policy}} = \left\| \hat{z}_t - z_t \right\|_2^2$, where $\hat{z}_t = \pi(o_t)$.

\noindent\textbf{Action Decoder.}
We implement the action decoder as a single-pass transformer decoder similar to prior work~\cite{zhao2023learning, liu2026oat}. The decoder is trained only on data with ground-truth action labels using an action reconstruction objective. Given a latent action $z_t$ encoded from frames $(o_t, o_{t+k})$, the action loss $\mathcal{L}_{\text{act}} = \left\| \hat{a}_{t:t+k} - a_{t:t+k} \right\|_2^2$, where $\hat{a}_{t:t+k} = \psi(z_t)$.

\subsection{Joint Training with Representation Alignment}

\begin{figure*}[t]
    \centering
    \includegraphics[width=1.0\textwidth]{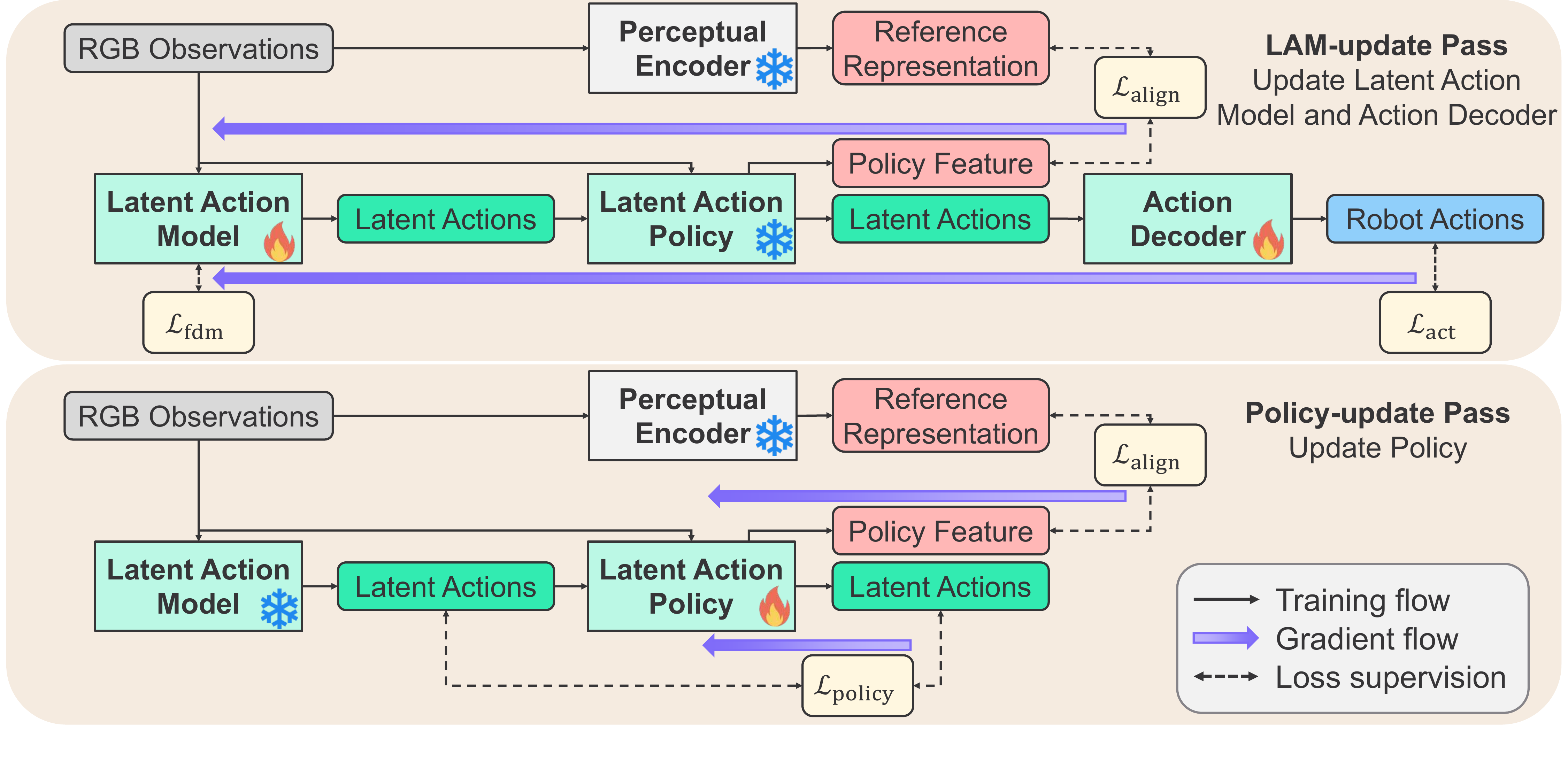}
    \vspace{-30pt}
    \caption{
        \textbf{End-to-End Joint Training.} \methodname{} jointly trains the latent action model (LAM), latent action policy, and action decoder using two alternating forward passes. In the \emph{LAM-update pass}, the policy is frozen and the alignment, action, and FDM losses are backpropagated to update the LAM. In the \emph{Policy-update pass}, the LAM is frozen and the alignment and policy losses are used to update the policy network.
    }
    \label{fig: end2end}
    \vspace{-15pt}
\end{figure*}

In our formulation of latent action policy learning, we jointly train the latent action model (LAM), policy, and action decoder so that the latent space is encouraged to exhibit both \emph{modelability} and \emph{decodability}. However, a naive end-to-end objective that combines all losses into a single optimization step leads to latent action collapse in the IDM, a failure mode observed in prior work~\cite{liang2025clam, vahdat2021score}. In this setting, policy gradients bias the latent space toward representations that are easy to predict, even when they conflict with the objectives required for accurate decoding and LAM learning.

To address this issue, instead of directly propagating policy gradients into the LAM, we align both models to a shared reference representation space. Given visual representations produced by a frozen perceptual encoder,~\eg DINO~\cite{cijo2025dinov3}, we adopt a representation alignment objective~\cite{yu2024representation, leng2025repa} that regularizes the latent space around semantically meaningful visual structure while preserving action decodability. The policy is trained to align with the same reference representation.

Specifically, we extract features from the intermediate layer of the policy encoder and align both the IDM and policy to a shared representation produced by the frozen perceptual encoder. To avoid shortcut problem arising from direct joint optimization, we alternate between two update passes, as illustrated in Figure~\ref{fig: end2end}.
In the \emph{LAM-update pass}, the policy network is frozen, and the alignment loss is backpropagated only into the LAM. Because the policy remains fixed, the IDM is encouraged to produce latent actions whose induced policy features align with the reference representation, reshaping the latent space toward targets that are easier for the policy to model.
In the \emph{Policy-update pass}, the LAM is frozen while the same alignment objective is used to optimize the policy. This objective pushes the policy features toward the shared reference space, improving representation quality and latent action prediction.
Together, the two passes improve modelability from complementary directions: the \emph{LAM-update pass} encourages latent actions that are easier to predict, while the \emph{Policy-update pass} improves the policy’s ability to predict them. Both are coordinated through a shared observation-derived representation space.

\noindent\textbf{Representation Alignment Loss.}
The alignment objective is defined as
\begin{equation}
    \mathcal{L}_{\mathrm{align}}(\phi,\theta)
    =
    -
    \mathbb{E}_{(o_t,o_{t+k})\sim\mathcal{D}}
    \Bigl[
        \mathrm{Sim}\!\left(
        \mathbf{y},
        h_{\theta}(\tilde{\mathbf{h}})
        \right)
    \Bigr],
    \label{eq:align}
\end{equation}
where $\mathbf{y}=\mathcal{F}(o_t,o_{t+k})$ denotes the reference representation extracted from a frozen perceptual encoder $\mathcal{F}$. The feature $\tilde{\mathbf{h}}$ is taken from an intermediate layer of the policy encoder. Specifically, let $g_{\theta}$ denote the mapping defined by the policy input embedding followed by the first $L$ transformer blocks, where $L$ corresponds to the midpoint of the transformer stack. Given latent action $z$ and observation $o_t$, we extract the intermediate policy feature $\tilde{\mathbf{h}} := g_{\theta}(z, o_t),$ which is then projected through a trainable MLP head $h_{\theta}$ before being aligned with the reference representation $\mathbf{y}$. Here, $\mathrm{Sim}(\cdot,\cdot)$ denotes cosine similarity between the reference representation $\mathbf{y}$ and the projected policy feature $h_{\theta}(\tilde{\mathbf{h}})$.

\noindent\textbf{Overall Training.}
We optimize the following objective:
\begin{align*}
    \mathcal{L}(\theta, \phi, \omega)
    =
    \mathcal{L}_{\mathrm{fdm}}(\theta)
    +
    \mathcal{L}_{\mathrm{act}}(\theta)
    +
    \mathcal{L}_{\mathrm{align}}(\theta, \phi, \omega)
    +
    \mathcal{L}_{\mathrm{policy}}(\phi),
\end{align*}
where $\theta$, $\phi$, and $\omega$ denote the parameters of the LAM, policy, and MLP projection head, respectively.
All losses are optimized over a mixture of action-free and action-labeled video data, except for $\mathcal{L}_{\mathrm{act}}$, which is applied only to action-labeled samples.

\noindent\textbf{Implementation Details.}
We use a frozen DINOv3~\cite{cijo2025dinov3} model as the perceptual encoder for computing the reference representations in the alignment objective.
We employ a warm-up stage of 5k iterations in which only $\mathcal{L}_{\mathrm{fdm}}$ and $\mathcal{L}_{\mathrm{act}}$ are optimized. The warm-up allows the IDM to first learn a stable latent action space before introducing alignment and policy learning.

\section{Experiments}

Our experiments are designed to investigate the following research questions.
\begin{enumerate}[noitemsep, topsep=0pt]
    \item \textbf{Few-shot Learning from Action-Free Video in Simulation} (Section~\ref{sec: sim_action_free}): Can \methodname{} improve task performance in simulation by leveraging additional robot video alongside a small set of action-labeled robot demonstrations, beyond methods trained only on action-labeled data?
    
    \item \textbf{Cross-Embodiment Learning in the Real World} (Section~\ref{sec: real_actino_free}): Can \methodname{} effectively learn real-world policies from a combination of human video demonstrations and a limited set of action-labeled robot demonstrations?

    \item \textbf{Analysis of Decodability and Modelability} (Section~\ref{sec: analysis} and~\ref{sec: linear_probe}): How does \methodname{} compare to IDM baselines trained with significantly larger data budgets in terms of decodability and modelability under few-shot settings?
    
\end{enumerate}

\subsection{Few-shot Policy Learning from Action-Free Data}
\label{sec: sim_action_free}

\noindent\textbf{Benchmark Setup.}
We evaluate all baselines on LIBERO-10, the most challenging subset of the LIBERO benchmark~\cite{liu2023libero}, which consists of 10 long-horizon tabletop manipulation tasks. For training of each task, we provide 50 robot demonstrations, of which $N$ samples are action-labeled, where $N \in \{5, 10, 20\}$. Task success is measured in simulation, and results are averaged over 3 runs with different random seeds.

\noindent\textbf{Baselines.}
To evaluate the effectiveness of our approach in a data-constrained setting with limited action-labeled data and abundant action-free data, we apply \methodname{} to two representative imitation learning baselines: Diffusion Policy (DP) and Unified World Model (UWM)~\cite{zhu2025unified}, which represent standard imitation learning and world-action modeling approaches, respectively. We extend both baselines with our IDM, action decoder, and end-to-end training recipe, resulting in $\text{\textsc{Ela}}_{\text{DP}}$ and $\text{\textsc{Ela}}_{\text{UWM}}$, respectively.
We further compare against state-of-the-art methods designed specifically for learning from video demonstrations, including ATM~\cite{wen2023any} and AMPLIFY~\cite{collins2025amplify}. Notably, both ATM and AMPLIFY require specialized data preprocessing pipelines that depend on external point-tracking models, whereas our approach does not rely on such additional components or preprocessing steps.

\noindent\textbf{Training Details.}
 DP cannot leverage action-free data and is therefore trained only on the $N$ action-labeled demonstrations available for each task. In contrast, all the other methods can utilize action-free data and are trained on a mixture of action-free and action-labeled data. We consider the single-task policy setting, in which all methods are first pretrained on all multi-task data (500 demonstrations) and then fine-tuned using 50 demonstrations from each target task. During \methodname{} fine-tuning, the IDM and Action Decoder remain frozen, allowing the policy to adapt only within the latent action manifold learned during pretraining.

\begin{figure}[t]
    \centering
    \includegraphics[width=0.9\textwidth]{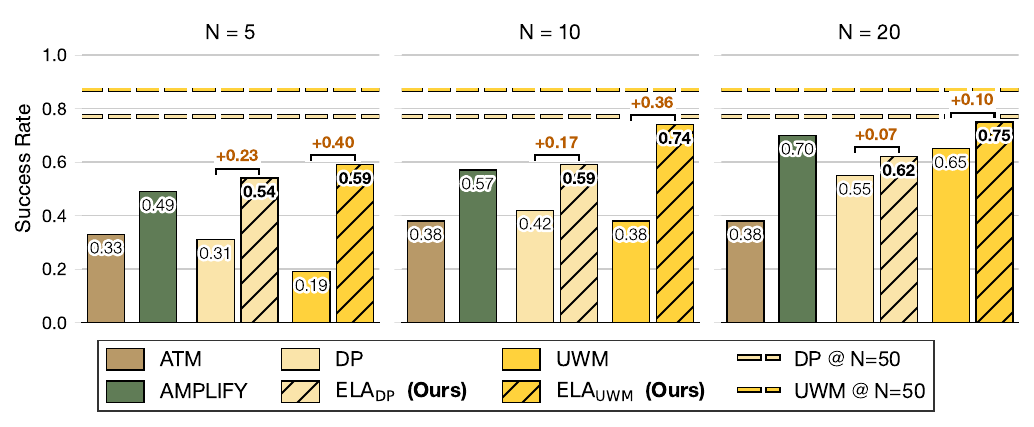}
    \vspace{-10pt}
    \caption{
        \textbf{Few-shot Policy Learning.} Comparative analysis of success rates on the LIBERO-10 benchmark. All evaluated methods were trained using $N$ action-labeled samples selected from 50 robot demonstrations per task. The two dashed lines at the top indicate the upper-bound performance of DP and UWM trained on the full set of 50 action-labeled demonstrations.
    }
    \vspace{-15pt}
    \label{fig: exp_libero}
\end{figure}

\noindent\textbf{Results.}
As shown in Figure~\ref{fig: exp_libero}, UWM outperforms DP when trained with the full set of 50 action-labeled demonstrations, corresponding to the upper-bound setting. However, in low-data regimes with only $N \in \{5,10\}$ action-labeled demonstrations, UWM underperforms DP, highlighting the greater difficulty of training world action models compared to standard imitation learning methods under limited supervision.
Despite this challenge, \methodname{} consistently improves the performance of both DP and UWM by enabling supervision through latent actions inferred from all available demonstrations, including both action-free and action-labeled samples. For instance, when N=5, \methodname{} improves UWM from $19\%$ to $59\%$ success rate, and DP from $31\%$ to $54\%$. In the low-data setting $N \in \{5,10\}$, both $\text{\textsc{Ela}}_{\text{DP}}$ and $\text{\textsc{Ela}}_{\text{UWM}}$ outperform prior state-of-the-art methods such as ATM and AMPLIFY, which rely on external models and specialized policy architectures.

\subsection{Cross-Embodiment Learning from Human Data}
\label{sec: real_actino_free}

\noindent\textbf{Benchmark Setup.}
To evaluate the effectiveness of our approach in the cross-embodiment learning setting, we consider scenarios with limited action-labeled robot data and more abundant human video data. For each task, we use 10 action-labeled robot demonstrations and 40 human demonstrations.
Our real-world task suite covers a range of everyday activities, from standard pick-and-place tasks to manipulation of articulated objects. Task details are provided in the supplementary material. For evaluation, object locations are randomly initialized within the region shared between the robot camera’s field of view and the robot arm’s reachable workspace. For each task, we run 20 trials and measure success.

\begin{wraptable}{r}{0.45\textwidth}
    \vspace{-10pt}
    \small
    \centering
    \begin{tabular}{lcc}
        \toprule
        Baseline & Model        & Training Data \\
                 & Param.   & (\# trajectories) \\
        \midrule
        LAPA~\cite{ye2025latent}          & 206M               & 1M+ \\
        Villa-X~\cite{chen2025villa}        & 124M               & 1M+ \\
        UniVLA~\cite{bu2025univla}        & 34M$^{\dagger}$    & 1M+ \\
        \midrule
        LAPA-DINO~\cite{nie2026lary}      & 102M$^{\dagger}$   & 250 \\
        \textbf{$\text{\textsc{Ela}}_{\text{DP}}$ (Ours)} & 6M            & 250 \\
        \bottomrule
    \end{tabular}

    \vspace{-5pt}
    \caption{
        IDM model sizes and training budgets for baselines. $^{\dagger}$ DINO encoder parameters are excluded from the count.
    }
    \vspace{-10pt}
    \label{tab: baselines}
\end{wraptable}

\noindent\textbf{Baselines.} 
We evaluate the effectiveness of learned latent actions in our policy learning setup by comparing against several baselines, including LAPA~\cite{ye2025latent}, Villa-X~\cite{chen2025villa}, and UniVLA~\cite{bu2025univla}. These methods are trained on large-scale cross-embodiment datasets containing both human and robot demonstrations. Their model sizes and training data budgets are summarized in Table~\ref{tab: baselines}. Note that the reported parameter counts include only the IDM component and exclude additional modules such as the FDM.
We further include LAPA-DINO, a variant of LAPA that incorporates DINOv3~\cite{cijo2025dinov3} features and achieves state-of-the-art performance on the LARY benchmark~\cite{nie2026lary}.
We denote our full method as $\text{\textsc{Ela}}_{\text{DP}}$, which augments DP with an IDM and an action decoder. We also include an ablation variant, $\text{\textsc{Ela}}_{\text{DP}}$ (w/o align), which removes the representation alignment module.

\noindent\textbf{Training Details.}
All baselines except \methodname{} follow a two-stage training pipeline, where the IDM is trained independently from the policy. We use publicly available IDM weights as frozen encoders and train a downstream policy and action decoder on top of them. For LAPA-DINO, we additionally train the model on our dataset to serve as an in-domain baseline, while other methods are used directly without further adaptation.

\begin{figure}[t]
    \centering

    \begin{subfigure}{0.32\textwidth}
        \centering
        \includegraphics[width=\linewidth]{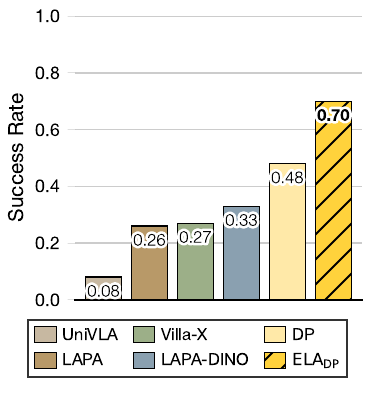}
        \vspace{-25pt}
        \caption{}
    \end{subfigure}
    \hfill
    \begin{subfigure}{0.32\textwidth}
        \centering
        \includegraphics[width=\linewidth]{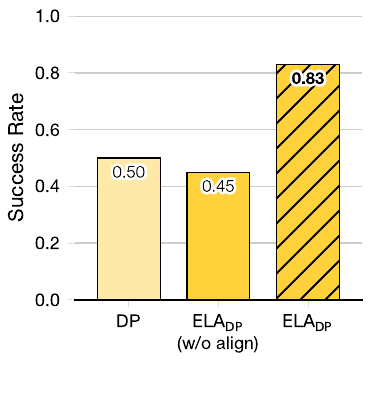}
        \vspace{-25pt}
        \caption{}
    \end{subfigure}
    \hfill
    \begin{subfigure}{0.32\textwidth}
        \centering
        \includegraphics[width=\linewidth]{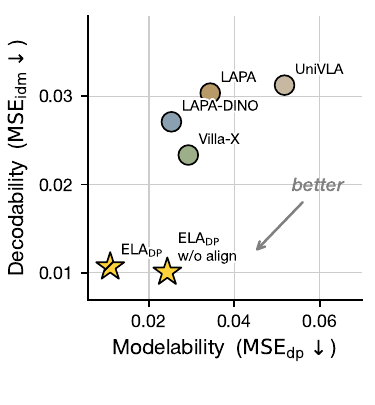}
        \vspace{-25pt}
        \caption{}
    \end{subfigure}

    \vspace{-5pt}
    \caption{
        \textbf{Real-world Results.} (a) Comparative analysis of success rates on the real-world benchmark. (b) Ablation study of representation alignment across a subset of task suites. (c) \emph{Modelability} versus \emph{Decodability} for each latent action spaces.
    }
    \vspace{-20pt}
    \label{fig: real}
\end{figure}

\noindent\textbf{Results.}
As shown in Figure~\ref{fig: real}(a), $\text{\textsc{Ela}}_{\text{DP}}$ achieves the best performance across all evaluated methods. Although LAPA, Villa-X, and UniVLA have substantially larger model capacities and are pre-trained on large-scale datasets, their gains do not translate into improved downstream policy performance in this setting. Meanwhile, despite being trained on in-domain demonstrations, LAPA-DINO still exhibits a substantial performance gap compared to $\text{\textsc{Ela}}_{\text{DP}}$ and even the baseline DP model. These results suggest that pretraining scale and data diversity alone are insufficient to ensure effective downstream policy learning in the few-shot regime.

To control for differences in data access, we additionally evaluate Dynamo~\cite{cui2024dynamo}, a self-supervised policy-pretraining method based on world-modeling objectives. Unlike standard DP, which cannot directly leverage human data for policy learning, Dynamo can exploit both human and robot videos. We pretrain Dynamo on videos collected across all tasks and then fine-tune the resulting representation for diffusion policy learning. As shown in Table~\ref{tab: rebuttal}, Dynamo underperforms both standard DP and $\text{\textsc{Ela}}_{\text{DP}}$, suggesting that access to human and robot data alone is insufficient to achieve strong downstream policy performance. One possible explanation is that image-representation pretraining is more susceptible to cross-embodiment appearance shifts, whereas $\text{\textsc{Ela}}_{\text{DP}}$ learns modelable and decodable latent actions that provide more direct supervision for cross-embodiment policy learning.

\begin{wraptable}{r}{0.30\textwidth}
    \vspace{-10pt}
    \small
    \centering
    \begin{tabular}{lc}
        \toprule
        Method & Success \\
        \midrule
        $\text{Villa-X}_{\text{FS}}$ & 0.06 \\
        $\text{Villa-X}_{\text{FT}}$ & 0.21 \\
        Dynamo & 0.38 \\
        DP & 0.48 \\
        \textbf{$\text{\textsc{Ela}}_{\text{DP}}$} & \textbf{0.70} \\
        \bottomrule
    \end{tabular}

    \vspace{-5pt}
    \caption{
        Real-world success under matched data access.
    }
    \vspace{-10pt}
    \label{tab: rebuttal}
\end{wraptable}

We next investigate whether the performance gap of existing latent-action approaches can be attributed to a lack of in-domain training. To this end, we evaluate two additional Villa-X variants: $\text{Villa-X}_{\text{FT}}$, fine-tuned on our in-domain data, and 
$\text{Villa-X}_{\text{FS}}$, trained from scratch using the same data as $\text{\textsc{Ela}}_{\text{DP}}$. As shown in Table~\ref{tab: rebuttal}, both variants underperform the original Villa-X, with success rates of 0.21 and 0.06, respectively, compared with 0.27 for the original model. This result suggests that simply adapting existing latent-action models to in-domain data is insufficient to achieve strong performance in the few-shot regime.

Finally, to assess the contribution of our representation alignment approach, we conduct an ablation study. As shown in Figure~\ref{fig: real}(b), removing the alignment component reduces the success rate from 0.83 to 0.45, demonstrating the importance of representation alignment for downstream policy learning.

\subsection{Analysis of Decodability and Modelability}
\label{sec: analysis}

\noindent\textbf{Study Setup.} To further analyze the gap in real-world experiments, we evaluate each IDM latent action space along two dimensions in Figure~\ref{fig: real}(c) using mean squared error (MSE) against ground-truth actions on 20 held-out real-world robot demonstrations from Section~\ref{sec: real_actino_free}. \emph{Decodability} ($\text{MSE}_\mathrm{idm}$) measures how well actions can be reconstructed from IDM latents extracted from frame transitions, indicating how much action-relevant information is preserved. \emph{Modelability} ($\text{MSE}_\mathrm{dp}$) measures reconstruction error from latents predicted by a diffusion policy conditioned on observations, indicating how predictable the latents are for the policy. Since policy-predicted latents approximate IDM latents, $\text{MSE}_\mathrm{idm}$ is a lower bound on $\text{MSE}_\mathrm{dp}$. A useful latent space must be both decodable and modelable, corresponding to the lower-left region of Figure~\ref{fig: real}(c).

\noindent\textbf{Results.} IDM methods trained without an action decoder,~\ie, UniVLA, LAPA, and LAPA-DINO, exhibit poor decodability and modelability. Villa-X slightly improves both metrics by aligning latent actions with underlying robot states and actions through its FDM objective. $\text{\textsc{Ela}}_{\text{DP}}$ (w/o align) achieves strong decodability but remains sub-optimal in modelability. Our full approach $\text{\textsc{Ela}}_{\text{DP}}$ achieves the best performance on both axes, highlighting the importance of joint training and representation alignment in learning effective latent action spaces.

\subsection{Linear Probing for Action Prediction}
\label{sec: linear_probe}

\noindent\textbf{Study Setup.} Although Figure~\ref{fig: real}(c) shows that \methodname{} achieves lower action reconstruction errors in both decodability and modelability, these metrics may partly reflect differences in the learned action decoders. To better disentangle representation quality from decoder performance, we conduct an additional linear probing experiment following a standard protocol in representation learning~\cite{nie2026lary}. Specifically, we freeze each learned latent representation and train an identical linear probe to predict target actions for 10 epochs, using a learning rate of $10^{-3}$ and a cosine learning-rate schedule. We evaluate the probes on 20 held-out samples across all tasks. By keeping the probe architecture and training protocol fixed across methods, this experiment provides a more direct comparison of the action-relevant information accessible in each representation while reducing the influence of decoder capacity.

\begin{wraptable}{r}{0.2\textwidth}
    \vspace{-10pt}
    \footnotesize
    \centering
    \setlength{\tabcolsep}{3pt}
    \begin{tabular}{lc}
        \toprule
        Method & MSE \\
        \midrule
        UniVLA        & 0.043 \\
        LAPA          & 0.042  \\
        Villa-X        & 0.024 \\
        \midrule
        $\text{Villa-X}_{\text{FS}}$        & 0.059 \\
        $\text{Villa-X}_{\text{FT}}$        & 0.029 \\
        LAPA-DINO      & 0.029 \\
        \textbf{$\text{\textsc{Ela}}_{\text{DP}}$} & 0.015 \\
        \bottomrule
    \end{tabular}

    \vspace{-5pt}
    \caption{
    Linear probing for action regression.
    }
    \vspace{-15pt}
    \label{tab: linear_probe}
\end{wraptable}

\noindent\textbf{Results.} As shown in Table~\ref{tab: linear_probe}, $\text{\textsc{Ela}}_{\text{DP}}$ achieves the lowest action prediction MSE of 0.015, outperforming all evaluated baselines. Among the original pretrained baselines, Villa-X achieves the second-lowest MSE of 0.024, followed by LAPA-DINO and the fine-tuned Villa-X variant, both with an MSE of 0.029. Notably, fine-tuning Villa-X on in-domain data does not improve its linear probing performance, while training from scratch yields a substantially higher MSE of 0.059. These results complement our decodability and modelability analysis, providing additional evidence that $\text{\textsc{Ela}}_{\text{DP}}$ learns representations that make action-relevant information more accessible to a simple predictor, rather than relying solely on the capacity of its learned action decoder to recover the underlying actions.

\section{Limitations}
\methodname{} effectively learns from video data, but has not yet been validated at large scale. Scaling to cross-embodiment datasets~\cite{yang2025egovla, khazatsky2024droid} and evaluating across diverse scenarios are natural directions for future work.
Second, \methodname{} operates directly on raw RGB pixels and does not explicitly leverage depth or 3D priors. Incorporating visual foundation models, such as monocular depth estimation~\cite{bochkovskiy2025depth}, or training with 3D supervision could help mitigate this limitation.
Finally, our evaluation excludes highly dexterous or precision-intensive tasks, leaving fine-grained manipulation as future work.

\section{Conslusion}
In this paper, we introduce \methodname{}, an end-to-end framework for learning from video demonstrations via latent actions, jointly optimized for policy learning and action decoding. \methodname{} consists of a latent action model, a latent-action policy, and a lightweight decoder, enabling effective use of large-scale action-free video with only a small set of robot demonstrations.
Simulated and real-world experiments show consistent improvements over prior methods in data-constrained settings. Our analysis further indicates that the learned latent actions enhance both \emph{modelability} and \emph{decodability}.
Overall, \methodname{} highlights the promise of leveraging abundant video data to improve imitation learning.

\section*{Acknowledgements}
This work has taken place in the Learning Agents Research Group (LARG) at the Artificial Intelligence Laboratory, The University
of Texas at Austin.  LARG research is supported in part by the National Science Foundation (FAIN-2019844, NRT-2125858, OIC-2535195),
the Office of Naval Research (N00014-24-1-2550), Army Research Office (W911NF-17-2-0181, W911NF-23-2-0004, W911NF-25-1-0065), Lockheed
Martin, Nuro, Lyda Hill Philanthropies, and Good Systems, a research grand challenge at the University of Texas at Austin. The views and
conclusions contained in this document are those of the authors alone. Peter Stone serves as the Chief Scientist of Sony Corporate Technology Center and receives financial compensation for that role.  The terms of this arrangement have been reviewed and approved by the University of Texas at Austin in accordance with its policy on objectivity in research.

\bibliography{example}  %

@article{torabi2018behavioral,
  title={Behavioral cloning from observation},
  author={Torabi, Faraz and Warnell, Garrett and Stone, Peter},
  journal={arXiv preprint arXiv:1805.01954},
  year={2018}
}

@inproceedings{schmidt2024learning,
  title={Learning to act without actions},
  author={Schmidt, Dominik and Jiang, Minqi},
  booktitle={International Conference on Learning Representations},
  volume={2024},
  pages={9379--9395},
  year={2024}
}

@inproceedings{ye2025latent,
  title={Latent action pretraining from videos},
  author={Ye, Seonghyeon and Jang, Joel and Jeon, Byeongguk and Joo, Se June and Yang, Jianwei and Peng, Baolin and Mandlekar, Ajay and Tan, Reuben and Chao, Yu-Wei and Lin, Bill Yuchen and others},
  booktitle={International Conference on Learning Representations},
  volume={2025},
  pages={28213--28239},
  year={2025}
}

@article{chen2025villa,
  title={{Villa-X}: enhancing latent action modeling in vision-language-action models},
  author={Chen, Xiaoyu and Wei, Hangxing and Zhang, Pushi and Zhang, Chuheng and Wang, Kaixin and Guo, Yanjiang and Yang, Rushuai and Wang, Yucen and Xiao, Xinquan and Zhao, Li and others},
  journal={arXiv preprint arXiv:2507.23682},
  year={2025}
}

@article{bu2025univla,
  title={{UniVLA}: Learning to act anywhere with task-centric latent actions},
  author={Bu, Qingwen and Yang, Yanting and Cai, Jisong and Gao, Shenyuan and Ren, Guanghui and Yao, Maoqing and Luo, Ping and Li, Hongyang},
  journal={arXiv preprint arXiv:2505.06111},
  year={2025}
}

@article{liang2025clam,
  title={{CLAM}: Continuous latent action models for robot learning from unlabeled demonstrations},
  author={Liang, Anthony and Czempin, Pavel and Hong, Matthew and Zhou, Yutai and Biyik, Erdem and Tu, Stephen},
  journal={arXiv preprint arXiv:2505.04999},
  year={2025}
}

@article{zhang2026clap,
  title={{CLAP}: Contrastive Latent Action Pretraining for Learning Vision-Language-Action Models from Human Videos},
  author={Zhang, Chubin and Wang, Jianan and Gao, Zifeng and Su, Yue and Dai, Tianru and Zhou, Cai and Lu, Jiwen and Tang, Yansong},
  journal={arXiv preprint arXiv:2601.04061},
  year={2026}
}

@article{routray2025vipra,
  title={{ViPRA}: Video Prediction for Robot Actions},
  author={Routray, Sandeep and Pan, Hengkai and Jain, Unnat and Bahl, Shikhar and Pathak, Deepak},
  journal={arXiv preprint arXiv:2511.07732},
  year={2025}
}

@article{nie2026lary,
  title={{LARY}: A Latent Action Representation Yielding Benchmark for Generalizable Vision-to-Action Alignment},
  author={Nie, Dujun and Chen, Fengjiao and Lv, Qi and Kuang, Jun and Li, Xiaoyu and Cao, Xuezhi and Cai, Xunliang},
  journal={arXiv preprint arXiv:2604.11689},
  year={2026}
}

@article{darcet2023vision,
  title={Vision transformers need registers. arXiv},
  author={Darcet, T and Oquab, M and Mairal, J and Bojanowski, P},
  journal={arXiv preprint arXiv:2309.16588},
  year={2023}
}

@article{zhao2023learning,
  title={Learning fine-grained bimanual manipulation with low-cost hardware},
  author={Zhao, Tony Z and Kumar, Vikash and Levine, Sergey and Finn, Chelsea},
  journal={arXiv preprint arXiv:2304.13705},
  year={2023}
}

@article{lin2026ifsq,
  title={{iFSQ}: Improving FSQ for Image Generation with 1 Line of Code},
  author={Lin, Bin and Li, Zongjian and Niu, Yuwei and Gong, Kaixiong and Ge, Yunyang and Lin, Yunlong and Zheng, Mingzhe and Zhang, JianWei and Yang, Miles and Zhong, Zhao and others},
  journal={arXiv preprint arXiv:2601.17124},
  year={2026}
}

@article{liu2026oat,
  title={{OAT}: Ordered Action Tokenization},
  author={Liu, Chaoqi and Han, Xiaoshen and Gao, Jiawei and Zhao, Yue and Chen, Haonan and Du, Yilun},
  journal={arXiv preprint arXiv:2602.04215},
  year={2026}
}

@article{yu2024representation,
  title={Representation alignment for generation: Training diffusion transformers is easier than you think},
  author={Yu, Sihyun and Kwak, Sangkyung and Jang, Huiwon and Jeong, Jongheon and Huang, Jonathan and Shin, Jinwoo and Xie, Saining},
  journal={arXiv preprint arXiv:2410.06940},
  year={2024}
}

@article{vahdat2021score,
  title={Score-based generative modeling in latent space},
  author={Vahdat, Arash and Kreis, Karsten and Kautz, Jan},
  journal={Advances in neural information processing systems},
  volume={34},
  pages={11287--11302},
  year={2021}
}

@inproceedings{leng2025repa,
  title={{REPA-E}: Unlocking vae for end-to-end tuning of latent diffusion transformers},
  author={Leng, Xingjian and Singh, Jaskirat and Hou, Yunzhong and Xing, Zhenchang and Xie, Saining and Zheng, Liang},
  booktitle={Proceedings of the IEEE/CVF International Conference on Computer Vision},
  pages={18262--18272},
  year={2025}
}

@article{zhu2025unified,
  title={Unified world models: Coupling video and action diffusion for pretraining on large robotic datasets},
  author={Zhu, Chuning and Yu, Raymond and Feng, Siyuan and Burchfiel, Benjamin and Shah, Paarth and Gupta, Abhishek},
  journal={arXiv preprint arXiv:2504.02792},
  year={2025}
}

@article{li2025unified,
  title={Unified video action model},
  author={Li, Shuang and Gao, Yihuai and Sadigh, Dorsa and Song, Shuran},
  journal={arXiv preprint arXiv:2503.00200},
  year={2025}
}

@article{pai2025mimic,
  title={mimic-video: Video-action models for generalizable robot control beyond vlas},
  author={Pai, Jonas and Achenbach, Liam and Montesinos, Victoriano and Forrai, Benedek and Mees, Oier and Nava, Elvis},
  journal={arXiv preprint arXiv:2512.15692},
  year={2025}
}

@article{su2025motion,
  title={Motion before action: Diffusing object motion as manipulation condition},
  author={Su, Yue and Zhan, Xinyu and Fang, Hongjie and Li, Yong-Lu and Lu, Cewu and Yang, Lixin},
  journal={IEEE Robotics and Automation Letters},
  year={2025},
  publisher={IEEE}
}

@inproceedings{o2024open,
  title={Open {X-E}mbodiment: Robotic Learning Datasets and {RT-X} Models},
  author={O’Neill, Abby and Rehman, Abdul and Maddukuri, Abhiram and Gupta, Abhishek and Padalkar, Abhishek and Lee, Abraham and Pooley, Acorn and Gupta, Agrim and Mandlekar, Ajay and Jain, Ajinkya and others},
  booktitle={2024 IEEE International Conference on Robotics and Automation (ICRA)},
  pages={6892--6903},
  year={2024},
  organization={IEEE}
}

@article{khazatsky2024droid,
  title={Droid: A large-scale in-the-wild robot manipulation dataset},
  author={Khazatsky, Alexander and Pertsch, Karl and Nair, Suraj and Balakrishna, Ashwin and Dasari, Sudeep and Karamcheti, Siddharth and Nasiriany, Soroush and Srirama, Mohan Kumar and Chen, Lawrence Yunliang and Ellis, Kirsty and others},
  journal={arXiv preprint arXiv:2403.12945},
  year={2024}
}

@article{yang2025egovla,
  title={{EgoVLA}: Learning vision-language-action models from egocentric human videos},
  author={Yang, Ruihan and Yu, Qinxi and Wu, Yecheng and Yan, Rui and Li, Borui and Cheng, An-Chieh and Zou, Xueyan and Fang, Yunhao and Cheng, Xuxin and Qiu, Ri-Zhao and others},
  journal={arXiv preprint arXiv:2507.12440},
  year={2025}
}

@article{patel2025robotic,
  title={Robotic Manipulation by Imitating Generated Videos Without Physical Demonstrations},
  author={Patel, Shivansh and Mohan, Shraddhaa and Mai, Hanlin and Jain, Unnat and Lazebnik, Svetlana and Li, Yunzhu},
  journal={arXiv preprint arXiv:2507.00990},
  year={2025}
}

@article{chen2025large,
  title={Large Video Planner Enables Generalizable Robot Control},
  author={Chen, Boyuan and Zhang, Tianyuan and Geng, Haoran and Song, Kiwhan and Zhang, Caiyi and Li, Peihao and Freeman, William T and Malik, Jitendra and Abbeel, Pieter and Tedrake, Russ and others},
  journal={arXiv preprint arXiv:2512.15840},
  year={2025}
}

@article{haldar2025point,
  title={Point policy: Unifying observations and actions with key points for robot manipulation},
  author={Haldar, Siddhant and Pinto, Lerrel},
  journal={arXiv preprint arXiv:2502.20391},
  year={2025}
}

@article{hong2025hand,
  title={HAND Me the Data: Fast Robot Adaptation via Hand Path Retrieval},
  author={Hong, Matthew and Liang, Anthony and Kim, Kevin and Rajaprakash, Harshitha and Thomason, Jesse and B{\i}y{\i}k, Erdem and Zhang, Jesse},
  journal={arXiv preprint arXiv:2505.20455},
  year={2025}
}

@article{wen2023any,
  title={Any-point trajectory modeling for policy learning},
  author={Wen, Chuan and Lin, Xingyu and So, John and Chen, Kai and Dou, Qi and Gao, Yang and Abbeel, Pieter},
  journal={arXiv preprint arXiv:2401.00025},
  year={2023}
}

@article{collins2025amplify,
  title={AMPLIFY: Actionless Motion Priors for Robot Learning from Videos},
  author={Collins, Jeremy A and Cheng, Lor{\'a}nd and Aneja, Kunal and Wilcox, Albert and Joffe, Benjamin and Garg, Animesh},
  journal={arXiv preprint arXiv:2506.14198},
  year={2025}
}

@inproceedings{hsu2025spot,
  title={{SPOT}: {SE (3)} pose trajectory diffusion for object-centric manipulation},
  author={Hsu, Cheng-Chun and Wen, Bowen and Xu, Jie and Narang, Yashraj and Wang, Xiaolong and Zhu, Yuke and Biswas, Joydeep and Birchfield, Stan},
  booktitle={2025 IEEE International Conference on Robotics and Automation (ICRA)},
  pages={4853--4860},
  year={2025},
  organization={IEEE}
}

@article{chi2025diffusion,
  title={Diffusion policy: Visuomotor policy learning via action diffusion},
  author={Chi, Cheng and Xu, Zhenjia and Feng, Siyuan and Cousineau, Eric and Du, Yilun and Burchfiel, Benjamin and Tedrake, Russ and Song, Shuran},
  journal={The International Journal of Robotics Research},
  volume={44},
  number={10-11},
  pages={1684--1704},
  year={2025},
  publisher={Sage Publications Sage UK: London, England}
}

@article{kim2025uniskill,
  title={UniSkill: Imitating Human Videos via Cross-Embodiment Skill Representations},
  author={Kim, Hanjung and Kang, Jaehyun and Kang, Hyolim and Cho, Meedeum and Kim, Seon Joo and Lee, Youngwoon},
  journal={arXiv preprint arXiv:2505.08787},
  year={2025}
}

@article{liu2023libero,
  title={{LIBERO}: Benchmarking knowledge transfer for lifelong robot learning},
  author={Liu, Bo and Zhu, Yifeng and Gao, Chongkai and Feng, Yihao and Liu, Qiang and Zhu, Yuke and Stone, Peter},
  journal={Advances in Neural Information Processing Systems},
  volume={36},
  pages={44776--44791},
  year={2023}
}

@article{cijo2025dinov3,
  title={{DINOv3}},
  author={Sim{\'e}oni, Oriane and Vo, Huy V and Seitzer, Maximilian and Baldassarre, Federico and Oquab, Maxime},
  journal={arXiv preprint arXiv:2508.10104},
  volume={2},
  number={4},
  pages={5},
  year={2025}
}

@inproceedings{bochkovskiy2025depth,
  title={Depth pro: Sharp monocular metric depth in less than a second},
  author={Bochkovskiy, Alexey and Delaunoy, Ama{\"e}l and Germain, Hugo and Santos, Marcel and Zhou, Yichao and Richter, Stephan and Koltun, Vladlen},
  booktitle={International Conference on Learning Representations},
  volume={2025},
  pages={75602--75637},
  year={2025}
}

@article{cui2024dynamo,
  title={Dynamo: In-domain dynamics pretraining for visuo-motor control},
  author={Cui, Zichen J and Pan, Hengkai and Iyer, Aadhithya and Haldar, Siddhant and Pinto, Lerrel},
  journal={Advances in Neural Information Processing Systems},
  year={2024}
}

\appendix
\clearpage

\section{Latent Action Ablations}

To evaluate the impact of latent-action design choices, we ablate our approach $\text{\textsc{Ela}}_{\text{DP}}$ on the \texttt{soup-cheese} and \texttt{book-caddy} tasks from the LIBERO-10 benchmark under the \(N=10\) setting. The result is shown in Table~\ref{tab: latent_ablation}.

\noindent\textbf{Number of Latent-Action Tokens.}
We vary the latent-action length \(k \in \{4, 8, 16, 32\}\) while keeping all other hyperparameters fixed. Performance improves from \(0.59\) at \(k=4\) to a peak of \(0.83\) at \(k=16\), and then decreases slightly to \(0.78\) at \(k=32\). These results indicate that increasing the number of latent-action tokens improves performance up to a moderate value, while further increasing the sequence length provides no additional benefit. We therefore use \(k=16\) in all experiments.

\noindent\textbf{Codebook Size.}
We evaluate four FSQ level configurations: \([5,5,5,5]\) (625 codes), \([7,7,7,7]\) (2,401 codes), \([9,9,9,9]\) (6,561 codes), and \([7,7,7,7,7]\) (16,807 codes). The \([9,9,9,9]\) configuration achieves the highest success rate (\(0.83\)). Performance is substantially lower with the smallest codebook (\(0.47\)), while increasing the codebook size beyond 6,561 codes yields little to no improvement. Based on these results, we adopt \([9,9,9,9]\) as the default configuration.

\begin{table}[t]
\centering
\begin{tabular}{@{}cc@{\hspace{2em}}ccc@{}}
\toprule
\multicolumn{2}{c}{(a) Num. of latent-action tokens $k$} & \multicolumn{3}{c}{(b) FSQ codebook} \\
\cmidrule(r){1-2} \cmidrule(l){3-5}
$k$ & Success & Levels & Codebook size & Success \\
\midrule
$4$            & 0.59 & $[5,5,5,5]$       & $625$   & 0.47 \\
$8$            & 0.76 & $[7,7,7,7]$       & $2401$  & 0.78 \\
$16^{\dagger}$ & 0.83 & $[9,9,9,9]^{\dagger}$ & $6561$ & 0.83 \\
$32$           & 0.78 & $[7,7,7,7,7]$   & $16807$ & 0.57 \\
\bottomrule
\end{tabular}
\vspace{5pt}
\caption{Ablations on the latent action. We vary the number of latent-action tokens (k) (left) and the FSQ codebook configuration (right), one factor at a time while keeping all other settings fixed. The default configuration is marked with ($\dagger$).}
\vspace{-20pt}
\label{tab: latent_ablation}
\end{table}

\section{Real-world Experiment Details}

\subsection{Hardware Details}
\begin{wrapfigure}{r}{0.45\textwidth}
    \centering
    \vspace{-10pt}
    \includegraphics[width=\linewidth]{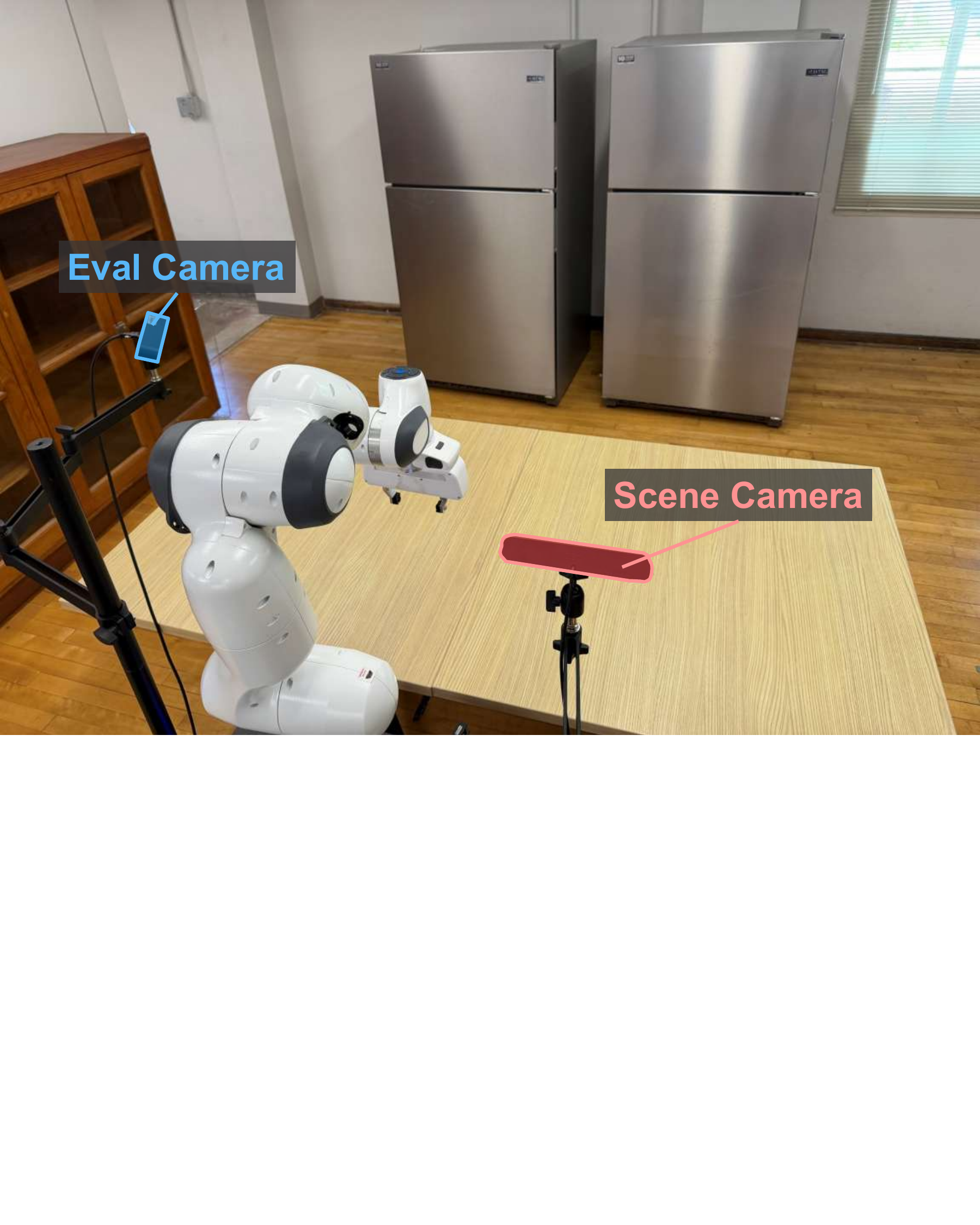}
    \vspace{-105pt}
    \caption{
        \textbf{Robot Setup.}
    }
    \label{fig: hardware}
    \vspace{-18pt}
\end{wrapfigure}
We conduct real-world experiments using a Franka Panda robot, as shown in Figure~\ref{fig: hardware}. The robot is mounted on a wheeled desk, allowing it to be easily deployed across a variety of environments and task scenarios. The observation space consists of a single scene camera, while an additional external camera is used solely for evaluation to monitor task initialization and execution.

The robot is controlled at 10 Hz using an Operational Space Control (OSC) controller. The action space is parameterized as a delta end-effector (EE) pose, consisting of incremental translational and rotational commands relative to the current EE pose. Gripper control is represented by a single continuous action dimension, where 0 denotes a fully open gripper and 1 denotes a fully closed gripper. This setup enables responsive and smooth execution of manipulation tasks.

\subsection{Baseline Details}
We compare against several latent action models. LAPA uses the publicly available checkpoint\footnote{\url{https://huggingface.co/latent-action-pretraining/LAPA-7B-openx}}, which is pretrained on the full Open X-Embodiment dataset~\cite{o2024open}. 
Villa-X employs the publicly released model\footnote{\url{https://huggingface.co/microsoft/villa-x}}. Since Villa-X is trained on a diverse collection of robot and egocentric human datasets, we refer readers to the original paper for a complete description of its training data.
For UniVLA, we use the latent action model\footnote{\url{https://huggingface.co/qwbu/univla-latent-action-model}} trained on Open X-Embodiment~\cite{o2024open} and Ego4D~\cite{grauman2021ego4d} dataset.
The implementation of LAPA-DINO is obtained from the LARYBench~\cite{nie2026lary} repository\footnote{\url{https://github.com/meituan-longcat/LARYBench}}. For a controlled comparison, we train LAPA-DINO using the same training dataset as our method, \methodname{}.

\subsection{Task Details}
We provide detailed descriptions of the tasks used in the real-world benchmark.
\begin{itemize}
    \item \texttt{place-food}: Pick up a bag of food from the counter and place it onto a plate.
    \item \texttt{place-food-cabinet}: Pick up a bag of food from the top cabinet and place it onto a plate on the counter. The task requires handling a larger vertical displacement and precise placement onto a constrained target region.
    \item \texttt{stack-bowl}: Grasp a bowl and stack it on top of another bowl. Successful execution requires precise alignment and stable object placement.
    \item \texttt{pour-bowl}: Pick up a container and perform a pouring motion into a target bowl. The task requires coordinated manipulation and accurate control of the object's orientation throughout the pouring trajectory.
    \item \texttt{open-oven}: Grasp the oven handle and open the oven door while respecting the articulation constraints of the door. Successful execution requires accurate handle grasping and coordinated motion along the door's opening trajectory.
\end{itemize}

\section{Detailed Results}
\begin{table}[t]
\centering
\begin{tabular}{lcccccc}
\toprule
{Task} &
{UniVLA} &
{LAPA} &
{Villa-X} &
{LAPA-DINO} &
{DP} &
$\mathbf{\text{\textsc{Ela}}_{\text{DP}}}$ \\
\midrule
place-food         & 0.25 & 0.55 & 0.50 & 0.50 & 0.60 & \textbf{0.95} \\
place-food-cabinet & 0.00 & 0.25 & 0.20 & 0.50 & 0.55 & \textbf{0.80} \\
stack-bowl         & 0.15 & 0.35 & 0.45 & 0.35 & 0.40 & \textbf{0.70} \\
pour-bowl          & 0.00 & 0.10 & 0.10 & 0.20 & 0.50 & \textbf{0.75} \\
open-oven          & 0.00 & 0.05 & 0.10 & 0.10 & \textbf{0.35} & 0.30 \\
\midrule
Average            & 0.08 & 0.26 & 0.27 & 0.33 & 0.48 & \textbf{0.70} \\
\bottomrule
\end{tabular}
\vspace{5pt}
\caption{Success rates on the real-world benchmark (20 trials per task).}
\label{tab: real}
\end{table}

\subsection{Real-world Benchmark}
Table~\ref{tab: real} reports the per-task success rates of all baselines on the real-world benchmark. Across all tasks, most baselines underperform the DP policy, primarily due to their reliance on learned inverse dynamics models (IDMs) that generate latent actions with limited decodability and modelability for downstream control. In contrast, $\text{\textsc{Ela}}_{\text{DP}}$ consistently improves over DP by incorporating additional human demonstrations and more effectively shaping the latent action space through our joint training recipe. The only exception is the \texttt{open-oven} task, where $\text{\textsc{Ela}}_{\text{DP}}$ does not provide further gains. This task demands highly precise continuous control under strong articulation constraints, which exceeds the modeling capacity of the IDM when learning from typical human demonstration videos, thereby limiting the benefit of additional data.

\newcommand{\mstd}[2]{$#1 \pm{\scriptstyle #2}$}

\begin{table*}[t]
\centering

\begin{tabular}{lcccc}
\toprule
{Task} &
{DP} &
$\text{\textsc{Ela}}_{\text{DP}}$ &
{UWM} &
$\text{\textsc{Ela}}_{\text{UWM}}$ \\
\midrule

\multicolumn{5}{l}{\textbf{N = 20}} \\

bowl-drawer &
\mstd{0.80}{0.04} &
\mstd{0.89}{0.08} &
\mstd{0.91}{0.06} &
\mstd{\mathbf{0.93}}{0.03} \\

cheese-butter &
\mstd{0.69}{0.08} &
\mstd{\mathbf{0.77}}{0.10} &
\mstd{0.75}{0.02} &
\mstd{0.74}{0.00} \\

moka-moka &
\mstd{0.43}{0.09} &
\mstd{0.26}{0.07} &
\mstd{0.38}{0.03} &
\mstd{\mathbf{0.49}}{0.03} \\

mug-microwave &
\mstd{0.58}{0.06} &
\mstd{0.85}{0.03} &
\mstd{0.66}{0.00} &
\mstd{\mathbf{0.91}}{0.01} \\

mug-mug &
\mstd{0.36}{0.04} &
\mstd{0.46}{0.03} &
\mstd{0.55}{0.06} &
\mstd{\mathbf{0.59}}{0.03} \\

mug-pudding &
\mstd{0.25}{0.15} &
\mstd{0.17}{0.04} &
\mstd{0.32}{0.07} &
\mstd{\mathbf{0.45}}{0.03} \\

soup-cheese &
\mstd{0.63}{0.11} &
\mstd{\mathbf{0.88}}{0.02} &
\mstd{0.71}{0.14} &
\mstd{0.83}{0.01} \\

soup-sauce &
\mstd{0.40}{0.04} &
\mstd{\mathbf{0.71}}{0.04} &
\mstd{0.64}{0.09} &
\mstd{0.65}{0.04} \\

stove-moka &
\mstd{0.42}{0.08} &
\mstd{0.57}{0.08} &
\mstd{0.81}{0.06} &
\mstd{\mathbf{0.95}}{0.01} \\

book-caddy &
\mstd{0.81}{0.05} &
\mstd{0.82}{0.05} &
\mstd{0.80}{0.00} &
\mstd{\mathbf{0.95}}{0.02} \\

\midrule
Average (N=20) &
0.55 &
0.62 &
0.65 &
\textbf{0.75} \\

\midrule

\multicolumn{5}{l}{\textbf{N = 10}} \\

bowl-drawer &
\mstd{0.71}{0.09} &
\mstd{\mathbf{0.93}}{0.02} &
\mstd{0.57}{0.13} &
\mstd{0.92}{0.04} \\

cheese-butter &
\mstd{0.27}{0.01} &
\mstd{0.43}{0.05} &
\mstd{0.47}{0.08} &
\mstd{\mathbf{0.70}}{0.03} \\

moka-moka &
\mstd{0.23}{0.05} &
\mstd{0.25}{0.08} &
\mstd{0.03}{0.02} &
\mstd{\mathbf{0.70}}{0.07} \\

mug-microwave &
\mstd{0.54}{0.09} &
\mstd{0.71}{0.06} &
\mstd{0.32}{0.07} &
\mstd{\mathbf{0.77}}{0.04} \\

mug-mug &
\mstd{0.28}{0.03} &
\mstd{\mathbf{0.47}}{0.02} &
\mstd{0.33}{0.01} &
\mstd{0.45}{0.03} \\

mug-pudding &
\mstd{0.14}{0.04} &
\mstd{0.23}{0.01} &
\mstd{0.22}{0.07} &
\mstd{\mathbf{0.33}}{0.06} \\

soup-cheese &
\mstd{0.39}{0.11} &
\mstd{0.65}{0.17} &
\mstd{0.39}{0.01} &
\mstd{\mathbf{0.79}}{0.02} \\

soup-sauce &
\mstd{0.31}{0.05} &
\mstd{0.54}{0.06} &
\mstd{0.42}{0.05} &
\mstd{\mathbf{0.83}}{0.07} \\

stove-moka &
\mstd{0.51}{0.02} &
\mstd{0.83}{0.11} &
\mstd{0.58}{0.14} &
\mstd{\mathbf{0.93}}{0.03} \\

book-caddy &
\mstd{0.80}{0.00} &
\mstd{0.85}{0.04} &
\mstd{0.49}{0.05} &
\mstd{\mathbf{0.95}}{0.02} \\

\midrule
Average (N=10) &
0.42 &
0.59 &
0.38 &
\textbf{0.74} \\

\midrule

\multicolumn{5}{l}{\textbf{N = 5}} \\

bowl-drawer &
\mstd{0.60}{0.09} &
\mstd{\mathbf{0.79}}{0.01} &
\mstd{0.36}{0.03} &
\mstd{\mathbf{0.81}}{0.01} \\

cheese-butter &
\mstd{0.25}{0.03} &
\mstd{0.58}{0.14} &
\mstd{0.11}{0.01} &
\mstd{\mathbf{0.69}}{0.06} \\

moka-moka &
\mstd{0.07}{0.05} &
\mstd{\mathbf{0.27}}{0.11} &
\mstd{0.03}{0.02} &
\mstd{0.16}{0.02} \\

mug-microwave &
\mstd{0.45}{0.01} &
\mstd{0.39}{0.02} &
\mstd{0.14}{0.03} &
\mstd{\mathbf{0.51}}{0.01} \\

mug-mug &
\mstd{0.23}{0.03} &
\mstd{0.47}{0.15} &
\mstd{0.14}{0.03} &
\mstd{\mathbf{0.53}}{0.06} \\

mug-pudding &
\mstd{0.02}{0.03} &
\mstd{0.06}{0.02} &
\mstd{0.07}{0.01} &
\mstd{\mathbf{0.19}}{0.06} \\

soup-cheese &
\mstd{0.41}{0.02} &
\mstd{0.37}{0.11} &
\mstd{0.05}{0.08} &
\mstd{\mathbf{0.62}}{0.04} \\

soup-sauce &
\mstd{0.17}{0.06} &
\mstd{0.47}{0.22} &
\mstd{0.14}{0.04} &
\mstd{\mathbf{0.63}}{0.03} \\

stove-moka &
\mstd{0.49}{0.08} &
\mstd{0.82}{0.21} &
\mstd{0.28}{0.10} &
\mstd{\mathbf{0.91}}{0.07} \\

book-caddy &
\mstd{0.41}{0.02} &
\mstd{0.63}{0.02} &
\mstd{0.56}{0.07} &
\mstd{\mathbf{0.84}}{0.00} \\

\midrule
Average (N=5) &
0.31 &
0.54 &
0.19 &
\textbf{0.59} \\

\bottomrule
\end{tabular}
\vspace{5pt}
\caption{LIBERO-10 results across different numbers of action-labeled demonstrations (N) out of 50 demonstrations per task. Each entry reports mean $\pm$ standard deviation over 3 runs. Best results per task are highlighted in bold.}
\label{tab: libero_multi_n}
\end{table*}

\subsection{Simulation Benchmark}
Table~\ref{tab: libero_multi_n} reports per-task success rates of all baselines on the LIBERO-10 benchmark. $\text{\textsc{Ela}}_{\text{DP}}$ and $\text{\textsc{Ela}}_{\text{UWM}}$ consistently improves both DP and UWM across all evaluated settings, demonstrating stable gains in the low-data regime. These results suggest that the benefits of $\text{\textsc{Ela}}$ are not confined to a specific data scale or model capacity, but instead generalize robustly across varying levels of data availability and task difficulty.

\section{Model and Training Details}
\begin{table}[t]
\centering
\small
\setlength{\tabcolsep}{6pt}
\renewcommand{\arraystretch}{1.15}
\begin{tabular}{l c c c}
\toprule
& \textbf{Phase 1} & \textbf{Phase 2} & \textbf{Phase 3} \\
& LAM Warmup & Pretraining & Per-task Finetuning \\
\midrule
Trained module(s)   & LAM $+$ Action Decoder     & LAM $+$ Action Decoder $+$ Policy & Policy \\
Frozen module(s)    & --             & --                    & LAM $+$ Action Decoder \\
\midrule
Steps               & $5\text{k}$  & $200\text{k}$         & $10\text{k}$ \\
Batch size          & $64$           & $64$                  & $64$ \\
Optimizer           & AdamW          & AdamW                 & AdamW \\
Learning rate       & $5\times10^{-5}$ & $5\times10^{-5}$    & $1\times10^{-4}$ \\
Weight decay        & $0$            & $0$                   & $1\times10^{-6}$ \\
$(\beta_1,\beta_2)$ & $(0.9,0.95)$   & $(0.9,0.95)$          & $(0.9,0.95)$ \\
Grad-norm clip      & $1.0$          & $1.0$                 & $1.0$ \\
LR schedule         & constant       & constant              & cosine ($1\text{k}$ warmup) \\
EMA                 & --             & --                    & \checkmark \\
\midrule
Losses              & $\mathcal{L}_{\mathrm{act}}$, $\mathcal{L}_{\mathrm{fdm}}$    & $\mathcal{L}_{\mathrm{act}}$, $\mathcal{L}_{\mathrm{fdm}}$, $\mathcal{L}_{\mathrm{policy}}$, $\mathcal{L}_{\mathrm{align}}$         & $\mathcal{L}_{\mathrm{policy}}$ \\
\bottomrule
\end{tabular}

\vspace{5pt}
\caption{\methodname{} Training schedule}

\label{tab: schedule}
\end{table}

\noindent\textbf{Inverse Dynamics Model.}
Our inverse dynamics model (IDM) is a register-based vision transformer that infers a compact latent action from an observation pair. Each frame is tokenized by a lightweight convolutional patch embedder. The transformer encoder then processes the frame pair with $16$ learnable register tokens summarize the visual context under a causal-last attention mask, inducing a coarse-to-fine ordering in which each register conditions only on its predecessors. The register tokens are projected to a $4$-dimensional latent and discretized by iFSQ.

\noindent\textbf{Forward Dynamics Model.}
To ground the latent space in physical dynamics, a forward dynamics model (FDM) provides auxiliary self-supervision: a 2-layer transformer decoder with learnable patch queries cross-attends to the first-frame tokens and the inferred registers, and a U-Net upsampling head reconstructs the target frame. First-frame feeding the FDM are detached so that the predictive signal shapes the latent action rather than leaking appearance information. The IDM and FDM are trained jointly end-to-end, with the action-prediction loss applied only to action-labeled samples and the FDM reconstruction loss applied to all pairs.

\noindent\textbf{Gripper Action Decoding.}
The action decoder maps latent actions to continuous robot control commands. Gripper control, however, is treated separately. This design is motivated by the nature of the latent action representation, which is learned through an inverse dynamics model trained on consecutive image pairs and therefore primarily captures visually observable state changes. While arm motions generate pronounced visual differences that are readily encoded in the latent action space, gripper open–close actions often result in only subtle, localized changes that are difficult to recover reliably from image observations alone.

To address this limitation, we exclude gripper control from the latent action space and instead predict it using a dedicated policy head supervised by the recorded binary gripper signal. Although this approach requires action labels for gripper prediction and therefore cannot directly leverage action-free demonstrations, our experiments show that separating gripper control from latent action decoding improves action reconstruction accuracy without compromising task performance in the low-data regime.

\noindent\textbf{Diffusion Policy.}
The downstream policy is formulated as a latent diffusion transformer. An ImageNet-pretrained ResNet-18 encodes the two-frame observation into a conditioning vector, while a 12-layer AdaLN Transformer denoises the $k \times (l_z{+}1)$ target, formed by fusing the $k \times l_z$ latent action with the gripper channel. Here, $k$ denotes the latent action token length (the number of register tokens) and $l_z$ the per-token latent dimension. The model is trained using a DDPM schedule, employing 50 diffusion steps during training and 10 steps during inference. Unless otherwise specified, we set $k = 16$ and $l_z = 4$ across all experiments.

\noindent\textbf{Representation Alignment.} 
Following REPA-E, we introduce a perceptual representation alignment objective that encourages the latent diffusion policy to match features from a frozen DINOv3 ViT-B/16 encoder. DINOv3 patch-token features are precomputed offline for all training clips. During training, hidden states from a fixed intermediate block of the diffusion transformer are projected through an MLP and aligned with the corresponding DINOv3 features using a negative cosine similarity loss.

\noindent\textbf{Training Schedule.}
\methodname{} training consists of three stages: a LAM warmup, a pretraining phase, and a fine-tuning phase. During pretraining, we jointly optimize the Latent Action Model (LAM), the latent action policy, and the action decoder to learn and organize the latent action space. In the fine-tuning stage, the IDM and Action Decoder are kept frozen, allowing the policy to adapt solely within the latent action manifold learned during pretraining. The complete training schedule and optimizer configurations for both stages are provided in Table~\ref{tab: schedule}.

\end{document}